\documentclass[10pt,twocolumn,letterpaper]{article}
\usepackage{cvpr}
\usepackage{times}
\usepackage{epsfig}
\usepackage{amsmath,amssymb}
\usepackage{amssymb}
\usepackage{graphicx}
\usepackage{amsmath,amssymb,amsthm}
\usepackage{bm}
\usepackage[linesnumbered,ruled]{algorithm2e}
\usepackage[american]{babel}
\newcommand{\T}{{\top}}
\newcommand{\st}{{\rm s.t.}}
\newcommand{\F}{{\rm F}}
\newcommand{\W}{{\bm A}}
\newcommand{\w}{{\bm a}}
\newcommand{\z}{{\bm z}}
\newcommand{\g}{{\bm m}}
\newcommand{\Z}{{\bm{Z}}}
\cvprfinalcopy
\usepackage[pagebackref=true,breaklinks=true,letterpaper=true,colorlinks,bookmarks=false]{hyperref}
\def\cvprPaperID{} 

\ifcvprfinal\fi
\begin{document}
\title{Generative Approach to Unsupervised Deep Local Learning}
\author{Changlu Chen$^1$, Chaoxi Niu$^1$, Xia Zhan$^2$, and Kun Zhan$^{1*}$\\
1. Lanzhou Univeristy, School of Information Science and Engineering\\
2. Qinghai Red Cross Hospital, Department of Statistics and Information\\
{\tt\small kzhan@lzu.edu.cn}
}

\maketitle
\begin{abstract}
Most existing feature learning methods optimize inflexible handcrafted features and the affinity matrix is constructed by shallow linear embedding methods. Different from these conventional methods, we pretrain a generative neural network by stacking convolutional autoencoders to learn the latent data representation and then construct an affinity graph with them as a prior. Based on the pretrained model and the constructed graph, we add a self-expressive layer to complete the generative model and then fine-tune it with a new loss function, including the reconstruction loss and a deliberately defined locality-preserving loss. The locality-preserving loss designed by the constructed affinity graph serves as prior to preserve the local structure during the fine-tuning stage, which in turn improves the quality of feature representation effectively. Furthermore, the self-expressive layer between the encoder and decoder is based on the assumption that each latent feature is a linear combination of other latent features, so the weighted combination coefficients of the self-expressive layer are used to construct a new refined affinity graph for representing the data structure. We conduct experiments on four datasets to demonstrate the superiority of the representation ability of our proposed model over the state-of-the-art methods.
\end{abstract}
\section{Introduction}
The success of learning algorithms depends highly on feature representation~\cite{Bengio6472238} and fully data-driven deep feature learning-based models have better performance than conventional handcrafted feature-based models due to powerful data representation ability of deep models~\cite{hinton2006reducing}. Meanwhile, unsupervised feature learning is a very important part of deep learning~\cite{Lecun2015Deep} since it is difficult to obtain an amount of high quality labeled data.

Unsupervised feature learning is one of the fundamental topics in the field of machine learning and computer vision. Subspace learning~\cite{vidal2011subspace}, being an especially important branch of unsupervised feature learning, aims to embed the low-level raw data into its latent space. In most subspace methods, each data point is represented by the combination of the whole data set and using the representation coefficients constructs a graph Laplacian for post-processing spectral clustering. Whether features can be well represented to explicitly reflect the data distribution turns out to be a critical factor to the success of unsupervised learning.
\begin{figure}[t]
  \centering
  \includegraphics[width=3.2in]{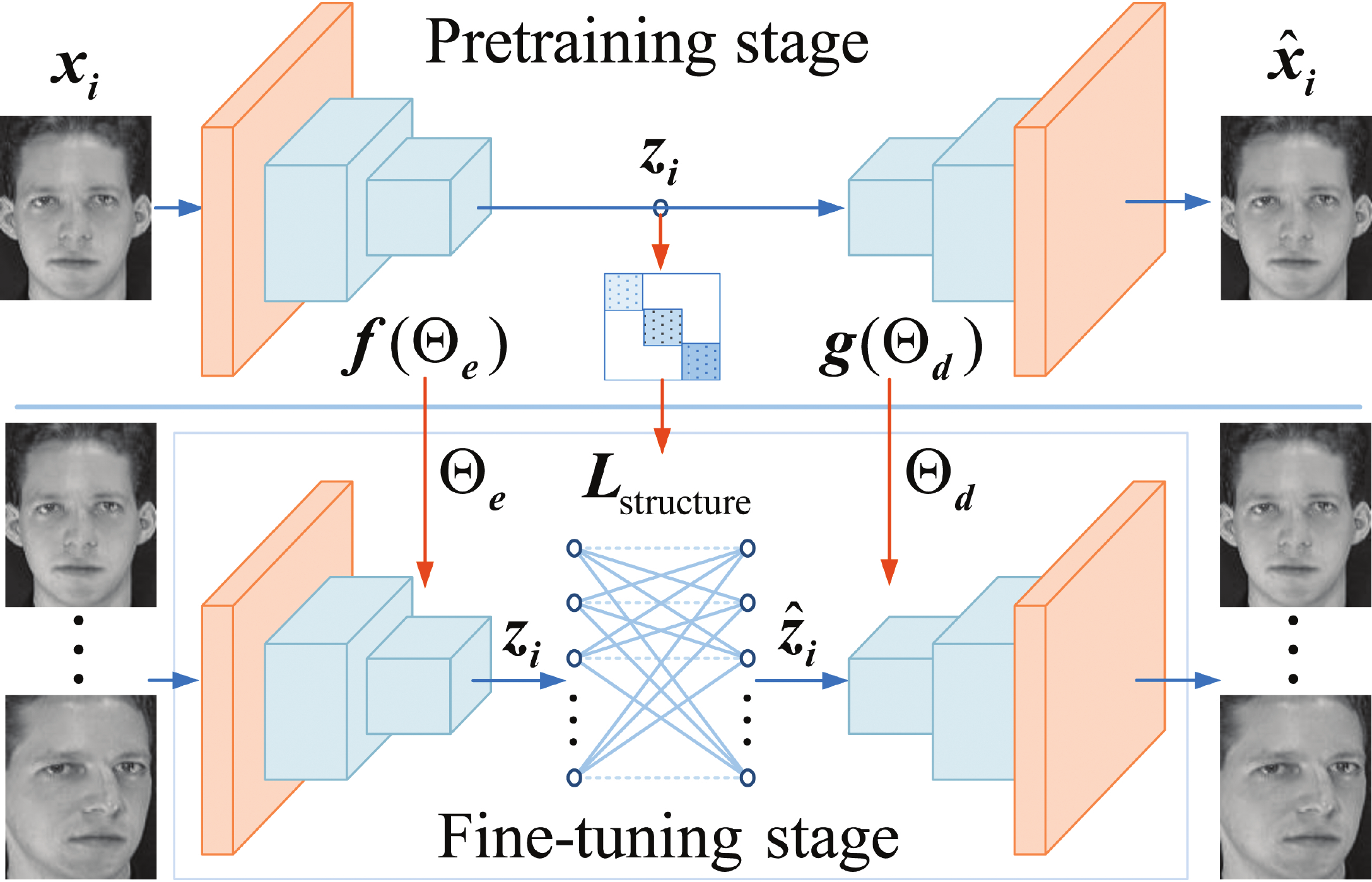}\\
  \caption{Unsupervised deep local learning. }\label{figure_01}
\end{figure}

Recently, many methods apply deep neural networks to unsupervised learning. Generally, these methods employ deep autoencoder generative model as an initialization, then the learned latent features are applied to different tasks. The loss function used in the fine-tuning stage consists of the network reconstruction error as well as affinity construction error with regularization. However, features obtained from these deep neural networks are directly fed into the fine-tuning stage without further exploiting local pairwise affinity of latent features, which inspires us to exploit well-distributed features with a locally-connected structure to significantly improve the quality of feature learning.

Most existing unsupervised learning methods suffer from some limitations. First, they use inflexible handcrafted features. Second, the representation affinity matrix is learned with shallow methods, such as sparse subspace learning~\cite{elhamifar2013sparse}, low rank representation~\cite{liu2013robust}, spectral curvature clustering~\cite{chen2009spectral} {\it etc}, which cannot adequately capture the latent data structure. Third, in order to exploit nonlinear functional relations from the raw data space to the latent feature space, the conventional methods use the kernel trick but they still remains a confusion in the choice of kernel function.

With the purpose to tackle the above challenges, as shown in Fig.~\ref{figure_01}, we frist pretrain a nonlinear generative neural network (GNN) model by stacking convolutional autoencoders, and then we construct an affinity graph with the pretrained GNN latent features to design a new locality-preserving loss function $\bm{L}_{\rm locality}$, and incorporate a self-expression layer in the fine-tuning stage.

In the proposed unsupervised deep local learning (UDLL) method, we focus on two common-sense facts and efficiently take into account the two facts in the well-designed model. Data points in the same cluster have strong connection with high similarity and a data point can be represented by others with coefficients weighing the pairwise affinity, which are the two common-sense facts.

The goal of normalized cut (Ncut) is to partition data points into $k$ weakly inter-connected and strongly intra-connected clusters~\cite{pamiNcut2000,ng2001spectral} where $k$ is the class number of objects, and Ncut can effectively reflect and preserve the raw data structure through predefining an affinity graph. Inspired by the great advantage of Ncut, we define a new locality-preserving loss to exploit the local connection information among the latent features. The locality-preserving loss renders latent space features to be of the intra-cluster compactness and inter-cluster separation during the fine-tuning stage. Thus, we use pretrained GNN latent features to construct an affinity graph and optimize with the locality-preserving loss in one integrated network, which can preserve connection structure from the pretrained latent features to the fine-tuned latent features.

Since each latent feature can be represented by other features, such a self-expressive layer is added in the middle of the encoder-decoder generative model and it is fine-tuned to learn a refined affinity between pairwise latent features. The weighted coefficients of the self-expressive layer reflect the pairwise affinity between latent features. With the locality-preserving loss, the structure of the affinity graph is gradually approximate to block-diagonal with reasonable connections during the model iteration.

Comparing to the conventional subspace feature learning methods, the proposed GNN-based deep feature learning method UDLL has following advantages:
\begin{enumerate}
  \item Since the handcrafted features can not well preserve important information from raw data, we use deep convolutional neural network features in this paper.
  \item We use pretrained deep GNN features to construct an affinity graph. Using the graph as prior knowledge, the locality-preserving loss is added to the loss function of UDLL. We use the locality-preserving loss besides reconstruction loss of the encoder-decoder GNN model in the fine-tuning stage to preserve local connection structure.
  \item In order to benefit from end-to-end optimization, we add a self-expressive layer in the middle of the encoder-decoder GNN model to learn a new refined affinity graph.
\end{enumerate}

The rest of the paper is organized as follows:
we introduce some typical deep neural networks-based feature learning methods in section \ref{RW}. Then, we introduce our UDLL algorithm in detail in section \ref{UDLL}. In section \ref{exp}, we demonstrate the efficiency of UDLL by conducting plentiful experiments and analyzing the results. Finally, we conclude our work in section \ref{cld}.
\section{Related Work}\label{RW}
Unsupervised learning with deep neural networks (DNN) is a relatively new topic. Autoencoders~\cite{hinton2006reducing,Vincent2010Stacked} is a typical DNN method to achieve the purpose of feature learning. In most of recent unsupervised DNN-based feature learning algorithms, autoencoders are used as a pretraining procedure to extract hierarchical data features.

\begin{table*}[!htbp]
  \caption{Comparison of the recently proposed unsupervised learning methods with our UDLL approach. CNN denotes convolutional neural network, SL denotes subspace learning, and LP denotes the locality preserving.}\label{comtabel}
  \centering
  \begin{tabular}{c|c|c|c|c|c|c}
    \hline
        &Tian~{\it et al.}\cite{tian2014learning}  &Ji~{\it et al.}\cite{ji2017deep}   &Xie~{\it et al.}\cite{xie2016unsupervised}&Guo~{\it et al.}\cite{Guo2017Improved} &Chang~{\it et al.}\cite{chang2017deep}&Ours   \\
    \hline
     CNN&$\times$  &\checkmark  &$\times$  &$\times$   &\checkmark &\checkmark   \\
     SL &$\times$  &\checkmark  &$\times$  &$\times$   &$\times$   &\checkmark   \\
     LP &$\times$  &$\times$    &$\times$  &\checkmark &$\times$   &\checkmark    \\
    \hline
  \end{tabular}
\end{table*}

Tian~{\it et al.}\cite{tian2014learning} use DNN to optimize the reconstruction loss function between the encoder and the decoder, but the input handcrafted feature is firstly optimized by subspace learning and then optimized by DNN secondly. Later, Peng~{\it et al.}~\cite{Peng2016Deep} also input handcrafted computer vision datasets for their DNN model resulting in that this method does not effectively utilize the representation ability of a convolutional neural network. As similar as Peng~{\it et al.}\cite{Peng2016Deep}'s method, Ji~{\it et al.}\cite{ji2017deep} also make use of the autoencoder as pretraining and self-expressive property to learn the affinity matrix. The subtle difference that Ji~{\it et al.}~\cite{ji2017deep}'s method inputs raw image data for a convolutional neural network rather than using handcrafted data in the Peng~{\it et al.}\cite{Peng2016Deep}'s model. Taking inspiration from $t$-SNE~\cite{maaten2008visualizing}, Xie~{\it et al.}\cite{xie2016unsupervised} define an centroid-based auxiliary target distribution to minimize Kullback-Leibler divergence, with parameters initialized by stacked autoencoders. Based on ~Xie~{\it et al.}'s\cite{xie2016unsupervised} algorithm, Dizaji~{\it et al.}\cite{dizaji2017deep} use cluster assignments frequency as a regularization term to balance the cluster results. Yang~{\it et al.}\cite{yang2016towards} jointly optimize a combination of the reconstruction error and $k$-means objective function to achieve `clustering-friendly' latent representations.

Inspired by the fact that deep convolutional neural networks can capture feature in a hierarchical way from a low-level to a high level, Chang~{\it et al.}\cite{chang2017deep} adopt the curriculum learning to adaptively select labeled samples for training convolutional neural networks and use a strategy to adaptively choose the label features defined by the cosine similarity. Yang~{\it et al.}\cite{Yang2016Joint} dispose of the successive clustering operations in a recurrent process, stacking the convolutional neural networks representations stepwise.
Guo~{\it et al.}\cite{Guo2017Improved} take the data structure into account, employing a clustering loss as prior to prevent the feature space from corruption. Tzoreff~{\it et al.}\cite{Tzoreff2018Deep} lay emphasis on the initial process of deep clustering and propose a discriminative pairwise loss function in terms of the autoencoder pretraining.
Based on the popular spectral clustering algorithm, Shaham~{\it et al.}\cite{Shaham2018SpectralNet} propose a deep neural network with a constraint in the last layer to satisfy the orthogonality property between the feature vectors.

We systematically compare our method with some of the related work in Table \ref{comtabel} to show the problem we solve. Among all of the above methods, we are the first to construct a graph by the pretraining features as prior knowledge of the fine-tuning stage. The local structure formed in the constructed prior graph is preserved from the pretrained latent feature to the fine-tuned latent feature. In the fine-tuning stage, we refine the latent feature to build a new affinity graph. We add the locality-preserving loss as a structure prior to fine-tune the block-diagonal affinity matrix with higher quality.
\section{Unsupervised Deep Local Learning}\label{UDLL}
\subsection{Generative Model}
Autoencoders are widely used in generative models and typically consist of an encoder and a decoder. As shown in Fig.~\ref{figure_01}, we adapt convolutional network to form the generative model, the parameters of the encoder are denoted by $\bm{\Theta}_e$, and decoder parameters are denoted by $\bm{\Theta}_d$. The encoder is denoted by a network $\bm{f}(\bm{\Theta}_e): \bm{x}_i\rightarrow \bm{z}_i$ and the decoder is $\bm{g}(\bm{\Theta}_d): \bm{z}_i\rightarrow \bm{\hat{x}_i}\,$. The loss function of the generative model is defined by the reconstruction cost,
\begin{equation}\label{ae}
\bm{L}(\bm{\Theta}_e,\bm{\Theta}_d)= \frac{1}{2}\|\bm{X} - \bm{\hat{X}}\|_\F^2
\end{equation}
where $\bm{X}=[\bm{x}_1,\bm{x}_2,\ldots,\bm{x}_n]$ is the input feature, $\bm{\hat{X}}=[\hat{\bm{x}}_1,\hat{\bm{x}}_2,\ldots,\hat{\bm{x}}_n]$ is the reconstruction feature, $n$ is the number of data points, and $\|\cdot\|_\F$ denotes the Frobenius norm.

\subsection{Prior Graph Construction}
We pretrain the generative model and use the pretrained feature $\bm{Z}=[\bm{z_1},\bm{z_2},\ldots,\bm{z_n}]$ to construct a graph.

Referring to the objective function of the normalized cut~\cite{pamiNcut2000,ng2001spectral}, we use the following objective to learn the affinity graph $\W=[a_{ij}]$,
\begin{equation}\label{CAN_sij}
\begin{split}
\min_{\W}&\sum_{i,j=1}^n\|\z_{i}-\z_{j}\|_2^2a_{ij}+ \lambda\|\W\|_\F^2\\
\st&~\W\geq0\,,~\W\bm{1}=\bm{1}\,.
\end{split}
\end{equation}
where $\lambda$ is a regularization parameter. By minimizing the above equation, if $\z_{i}$ and $\z_{j}$ have the similar feature, $a_{ij}$ would have a lager value which measure the similarity between them, and vice versa.

In Eq.~\eqref{CAN_sij}, we constrain $\W\bm{1}=\bm{1}$ so that $\W$ is a normalized graph and its degree matrix $\bm{D}$ is an identity matrix~\cite{pamiNcut2000} $\bm{D}=\bm{I}$, {\it i.e.}, $d_{jj}=\bm{1}^{\T}\w_{j} = 1$ where $\w_{j}$ is the $j$-th column of $\W$. We add the $\ell_2$-norm to smooth $\W$ otherwise the solution of Eq.~\eqref{CAN_sij} has trivial solution, {\it i.e.}, only one element is assigned to a value and others are zeroed.

Each column $\w_{j}$ of $\W$ is independent, so we can solve the following problem individually for each $j$:
\begin{equation}\label{OBJ_Ini_s}
\begin{split}
&\min_{\w_{j}}\sum_{i=1}^n \|\z_i-\z_j\|_2^2a_{ij}+ \lambda\sum_{i=1}^na_{ij}^2\\
&~~\st~\w_{j}\geq0\,,~{\bm 1}^\T\w_{j}=1\,.
\end{split}
\end{equation}

When we solve the $j$-th column $\w_j$, $\z_j$ is a fixed vector with respect to $\w_j$. Therefore, we can denote $\|\z_i-\z_j\|_2^2$ by a distance metric $m_{ij}$, and solving Eq.~\eqref{OBJ_Ini_s} is equal to optimizing the problem:
\begin{equation}\label{OBJ_sij}
\begin{split}
\min_{\w_j}& \frac{1}{2}\Bigl\|\w_j + \frac{1}{2\lambda}\g\Bigr\|_2^2\\
~~\st&~\w_{j}\geq0\,,~\bm{1}^{\T}\w_{j} = 1\,.
\end{split}
\end{equation}
where $\g=[m_{1j},m_{2j},\ldots,m_{nj}]^\T$ is a constant vector.

The Lagrangian function of Eq.~\eqref{OBJ_sij} is
\begin{equation}\label{L_sij}
\begin{split}
\mathcal{L}\left(\w_j\,,\eta\,,\bm{\rho}\right) =&\Bigl\|\w_j + \frac{1}{2\lambda}\g\Bigr\|_2^2 \\
 &- \eta (\bm{1}^{\T}\w_{j} - 1)- \bm{\rho}^{\T}\w_j
\end{split}
\end{equation}
where $\eta$ and $\bm{\rho}$ are the Lagrangian multipliers.

According to the Karush-Kuhn-Tucker condition \cite{boyd2004convex}, we have following equations,
\begin{eqnarray}
\w_j^{\star}&\geq&0\,;\label{KKT1}\\
\bm{1}^{\T}\w_{j}^{\star}&=&1\,;\label{KKT2} \\
\bm{\rho}^{\star}&\geq&0\,;\label{KKT3}\\
 \rho_i^{\star}a_{ij}^{\star}&=& 0\,,\forall\,i\in\{1,2,\ldots,n\}\,;\label{KKT4}\\
a_{ij}^{\star}+\frac{m_{ij}}{2\lambda}-\rho_i^{\star}- \eta^{\star}&=&0\,,\forall\,i\in\{1,2,\ldots,n\}\,.\label{KKT5}
\end{eqnarray}

Substituting Eq.~\eqref{KKT5} into Eq.~\eqref{KKT3}, we have
\begin{equation}\label{KKT6}
\eta^{\star}\leq a_{ij}^{\star}+\frac{m_{ij}}{2\lambda}\,,\forall\,i\in\{1,2,\ldots,n\}\,.
\end{equation}

Substituting Eq.~\eqref{KKT5} into Eq.~\eqref{KKT4}, we have
\begin{equation}\label{KKT7}
a_{ij}^{\star}(a_{ij}^{\star}+\frac{m_{ij}}{2\lambda}-\eta^{\star})=0\,,\forall\,i\in\{1,2,\ldots,n\}\,.
\end{equation}

If $\eta^{\star}>\frac{m_{ij}}{2\lambda}\,$, according to Eq.~\eqref{KKT6}, we have $a_{ij}^{\star}\geq\eta^{\star} -\frac{m_{ij}}{2\lambda}>0\,$. When $a_{ij}^{\star}>0\,$, the condition Eq.~\eqref{KKT7} can only hold if $a_{ij}^{\star} = -\frac{m_{ij}}{2\lambda} + \eta^{\star}\,$.

If $\eta^{\star}\leq\frac{m_{ij}}{2\lambda}\,$, then $a_{ij}^{\star}>0$ is impossible, because it would imply $a_{ij}^{\star}+\frac{m_{ij}}{2\lambda}-\eta^{\star}\geq a_{ij}^{\star}>0\,$, which violates Eq.~\eqref{KKT7}. Therefore, $a_{ij}^{\star}=0$ if $\eta^{\star}\leq\frac{m_{ij}}{2\lambda}\,$.

Thus, we have,
\begin{equation}\label{s_case}
a_{ij}^{\star}=
\begin{cases}
-\frac{m_{ij}}{2\lambda} + \eta^{\star}\,,&{\rm if}~\eta^{\star}>\frac{m_{ij}}{2\lambda}\,;\\
~~~~~~~~0\,,&{\rm if}~\eta^{\star}\leq\frac{m_{ij}}{2\lambda}\,.
\end{cases}
\end{equation}


Without loss of generality, we suppose that $\{m_{1j},$ $m_{2j},$ $\ldots,$ $m_{nj}\}$ are ordered from small to large. If there are only $k$ number of non-zero elements in $\w_j^{\star}\,$, then according to Eq.~\eqref{s_case}, we have $a^{\star}_{kj}>0$ and $a^{\star}_{k+1,j}=0\,$. Accoding to Eq.~\eqref{KKT2}, we have,
\begin{equation}\label{eta}
\sum_{i=1}^k\left(-\frac{m_{ij}}{2\lambda} + \eta^{\star}\right)=1\Rightarrow\eta^{\star} =\frac{2\lambda+\sum_{i=1}^km_{ij}}{2k\lambda}\,.
\end{equation}

Substituting $\eta^{\star}$ of Eq.~\eqref{eta} into Eq.~\eqref{s_case} and considering $a^{\star}_{kj}>0$ and $a^{\star}_{k+1,j}=0\,$, we have
\begin{equation}\label{beta}
\frac{k}{2}m_{kj}-\frac{1}{2}\sum_{i=1}^km_{ij} <\lambda\leq \frac{k}{2}m_{k+1,j}-\frac{1}{2}\sum_{i=1}^km_{ij}\,.
\end{equation}

In order to satisfy to condition in Eq.~\eqref{beta}, we set $\lambda$ to,
\begin{equation}\label{Beta}
\lambda= \frac{k}{2}m_{k+1,j}-\frac{1}{2}\sum_{i=1}^km_{ij}\,.
\end{equation}

According to Eq.~\eqref{Beta}, we have $2\lambda+\sum_{i=1}^km_{ij}=km_{k+1,j}\,$. Substituting it into Eq.~\eqref{eta}, we have $\eta^{\star}=\frac{m_{k+1,j}}{2\lambda}\,$. Thus, using $\eta^{\star}=\frac{m_{k+1,j}}{2\lambda}$ instead of Eq.~\eqref{s_case} and considering Eq.~\eqref{Beta}, the optimal affinity $a^{\star}_{ij}$ is a local $k$-nearest neighbor graph~\cite{Nie2016The},
\begin{equation}\label{s_star}
\w_{ij}^{\star}=
\begin{cases}
\frac{m_{k+1,j}-m_{ij}}{km_{k+1,j}-\sum_{p=1}^km_{pj}}\,,&{\rm if}~i\leq k\,;\\
~~~~~~~~0\,,&{\rm otherwise}\,.
\end{cases}
\end{equation}

The affinity matrix learned by Eq~\eqref{s_star} has many advantages: 1) naturally normalized because of $\bm{1}^\T\w_j=1$; 2) naturally sparse because sparseness is determined by the parameter $k$; and 3) its calculation only involves addition, subtraction, multiplication, and division~\cite{Nie2016The}. Thus, we use it in this paper.
\subsection{Unsupervised Feature Learning}
As shown in Fig.~\ref{figure_01}, we add a fully-connected layer called self-expressive layer in the middle of model between encoders and decoders. The UDLL network can achieve a nonlinear map from raw space to latent space, with the self-express layer to further learn a refined affinity matrix. We omit the biases and activations of self-expressive layer, and take $\Z = [\z_1,\z_2,\ldots,\z_n]$ as input and $\hat{\Z} = [\hat{\z}_1,\hat{\z}_2,\ldots,\hat{\z}_n]$ as output of the self-expressive layer. In the self-expressive layer, each feature $\hat{\z}_i$ can be represented with all other feature $\z_j$, {\it i.e.}, $\hat{\z}_i=\sum_{j=1}^nw_{ij}\z_{j}$ where $w_{ij}$ is weight of the layer. $w_{ij}$ has a natural meaning to subspace learning which can be thought as a weighting coefficient for representing affinity between $\hat{\z}_i$ and $\z_j$. With the self-expressive layer, we optimize the following overall loss in the fine-tuning stage,
\begin{equation}\label{obj}
\begin{split}
\bm{L}(\bm{\Theta}_e,\bm{W},\bm{\Theta}_d)=&\underbrace{\frac{1}{2}\|\bm{X} - \hat{\bm{X}}\|_\F^2}_{\bm{L}_{\rm reconstruction}} + \underbrace{\alpha \|\Z - \Z\bm{W}\|_\F^2 +\beta\|\bm{W}\|_\F^2}_{\bm{L}_{\rm affinity}} \\
&+\underbrace{\gamma\sum_{i,j=1}^n\|\z_{i}-\z_{j}\|_2^2a_{ij}}_{\bm{L}_{\rm locality}}
\end{split}
\end{equation}
where $\alpha$, $\beta$, and $\gamma$ are trade-off parameters.

In the overall loss function Eq.~\eqref{obj}, the first term is the reconstruction loss $\bm{L}_{\rm reconstruction}$, the second term is self-expressive loss with regularization $\bm{L}_{\rm affinity}$, and the third term is the locality-preserving loss $\bm{L}_{\rm locality}$ for preserving local structure from pretrained GNN feature space to fine-tuned feature space.

The proposed UDLL model can learn to output an affinity matrix $\bm{W}$ based on self-expressive property. The self-expressive property is inspired by conventional subspace learning methods~\cite{elhamifar2013sparse,liu2013robust,patel2014kernel,Xiao2016Robust}. By adding self-expressive layer, we can obtain a new refined affinity matrix $\bm{W}$ directly through fine-tuning the whole UDLL model. In this paper, the self-expressive loss with regularization is given by,
\begin{equation}\label{self-express}
\bm{L}_{\rm affinity}=\frac{1}{2}\|\bm{Z}-\bm{Z}\bm{W}\|^2_{\F} + \beta\|\bm{W}\|_\F^2\,.
\end{equation}

We use the pretrained deep GNN features to construct a prior graph $\W$ which depicts the affinity $a_{ij}$ between pretrained pairwise features $\z_i$ and $\z_j$. Then, we take into consideration the local connectivity between the pretrained latent data points to design the locality-preserving loss function. Since Ncut~\cite{pamiNcut2000,ng2001spectral} defines that vertices with strong connections are partitioned into one component and the weakly connected edges are cut off, inspired by Ncut, the locality-preserving loss is defined by
\begin{equation}\label{ncut}
\bm{L}_{\rm locality}=\sum_{i,j=1}^n\|\z_{i}-\z_{j}\|_2^2a_{ij}\,.
\end{equation}
Since the prior graph $\W$ is normalized by the constraint $\W\bm{1}=\bm{1}$, Eq.~\eqref{ncut} has the same form of original Ncut objective.

The detailed algorithm is summarized in Algorithm~\ref{algo}.
\begin{algorithm}
\caption{Unsupervised deep local learning.}\label{algo}
\SetKwInOut{Input}{input}
\SetKwInOut{Output}{output}
\SetKwInOut{Initialize}{initialize}
\Input{A dataset $\bm{X}$, the cluster number $k$, the epoch number $T$, $\alpha$, $\beta$, and $\gamma$.}
\Output{The fine-tuned latent feature $\bm{Z}$ and the refined affinity matrix $\bm{W}$.}
\Initialize{Pretrain the UDLL network without the self-expressive layer to learn $\bm{\Theta}_e$ and $\bm{\Theta}_d$ by optimizing Eq.~\eqref{ae}. Construct the affinity graph with the predefined features by Eq.~\eqref{s_star}.}
\For{$t\in\{1,2,\ldots,T\}$}
{
    Build the UDLL network with the additional self-expressive layer\;
    Fine-tune the network by optimizing Eq.~\eqref{obj} with back-propagation.
}
\end{algorithm}

In order to learn the new refined affinity graph $\bm{W}$ by the generative UDLL model, we add a self-expressive layer between encoders and decoders.
We take $\Z$ as input to the self-expressive layer, the weights of this layer correspond to the new affinity graph $\bm{W}$. The multiplication $\Z\bm{W}$ represents the newly combined features, and the difference between $\Z$ and $\Z\bm{W}$ measures the self-expressive error. With all of the above, the  new affinity graph $\bm{W}$ can be directly solved by Eq.~\eqref{obj}.


\subsection{Network Architecture}
As shown in Fig.~\ref{figure_01}, the UDLL network consists of three parts: an encoder, a self-expressive layer, and a decoder. The convolutional neural network is employed to build the encoder and the decoder. We use kernels with stride 2 in both horizontal and vertical directions and use rectified linear unit (ReLU)~\cite{Krizhevsky2012ImageNet} for nonlinear activations. By considering the connectivity of data points, the learned latent representation $\Z$ is more similar to the data points with the same label and more dissimilar to different label points, thus improving the quality of $\bm{W}$.

\begin{figure*}[ht]
\centering
\includegraphics[width=0.76\textwidth]{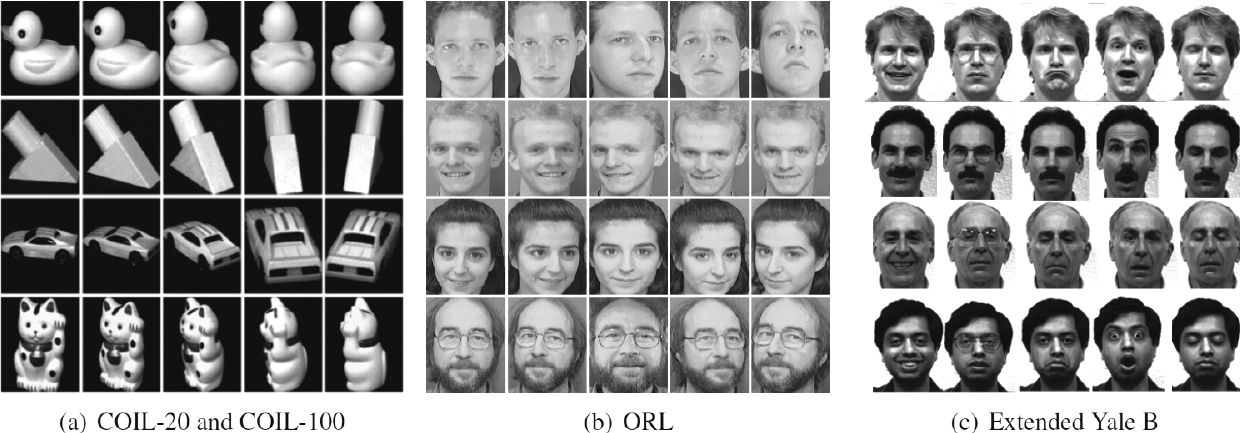}
\caption{Example images of different datasets.}\label{Example}
\end{figure*}

Suppose that the UDLL network has $l$-layer encoders and decoders with $\{c_1,c_2,\ldots, c_l\}$ channels. For the $i$-th encoder layer, if the kernel size is $s_i\times s_i$, the number of weights is $s_i^2c_i c_{i-1}$ with $c_0=1$. Due to the symmetric structure of autoencoders, the total number of weights is $\sum_{i = 1}^l 2s_i^2c_i c_{i-1}$ and the number of bias is $\sum_{i=1}^l 2c_i - c_1 + 1$. For the self-expressive layer, the number of $w_{ij}$ is $n^2$ when given $n$ number of input raw images. Thus, the total number of parameters of the whole UDLL network is:
\begin{equation}\label{com}
  \sum_{i=1}^l 2c_i(s_i^2 c_{i-1} + 1) - c_1 + 1 + n^2\,.
\end{equation}
\section{Experiments}\label{exp}
\subsection{Datasets}
We use four datasets in our experiment including, COIL-20~\cite{Nene1996}, COIL-100~\cite{Nayar1996}, ORL~\cite{Samaria1994}, and Extended Yale B~\cite{Lee2005Acquiring}. The dataset description is summarized in Table.~\ref{datasts}. Some sample images of these datasets is shown in Fig.~\ref{Example}.\\
{\bf COIL-20}~\cite{Nene1996} is from the Columbia object image library and contains 1440 images of 20 objects. Each object contains 72 images. Following~Cai~{\it et al.}\cite{Cai2011}, images are downsampled to $32 \times 32$.\\
{\bf COIL-100}~\cite{Nayar1996} is from the Columbia object image library and contains 7200 images of 100 different objects. Each object contains 72 images. Following~Cai~{\it et al.}\cite{Cai2011}, images are downsampled to $32 \times 32$.\\
{\bf ORL}~\cite{Samaria1994} contains 400 images of 40 distinct human faces and each subjects has 10 different images. Following Cai~{\it et al.}\cite{Cai2007}, original images are downsampled to $32 \times 32$. \\
{\bf Extended Yale B}~\cite{Lee2005Acquiring} contains 2432 facial images of 38 subjects which is represented by 64 images per subjects. These images are acquired under different illumination conditions. Following~Elhamifar~{\it et al.}\cite{elhamifar2013sparse}, images are downsampled to $42\times42$.
\begin{table}[!htbp]
  \centering
    \caption{Dataset description.}\label{datasts}
  \begin{tabular}{c|ccc}
     \hline
     Dataset & \# Image & \# Class & Image size \\
     \hline
     COIL-20 & 1440 & 20 & $32\times 32$ \\
     COIL-100 & 7200 & 100 & $32\times 32$ \\
     ORL & 400 & 40 & $32\times 32$ \\
     Yale & 2432 & 38 & $42\times 42$ \\
     \hline
  \end{tabular}
\end{table}
\subsection{Network Setting}
For different datasets, we use different convolutional neural network architectures.
\begin{figure*}[!ht]
\centering
\includegraphics[width=4.2in]{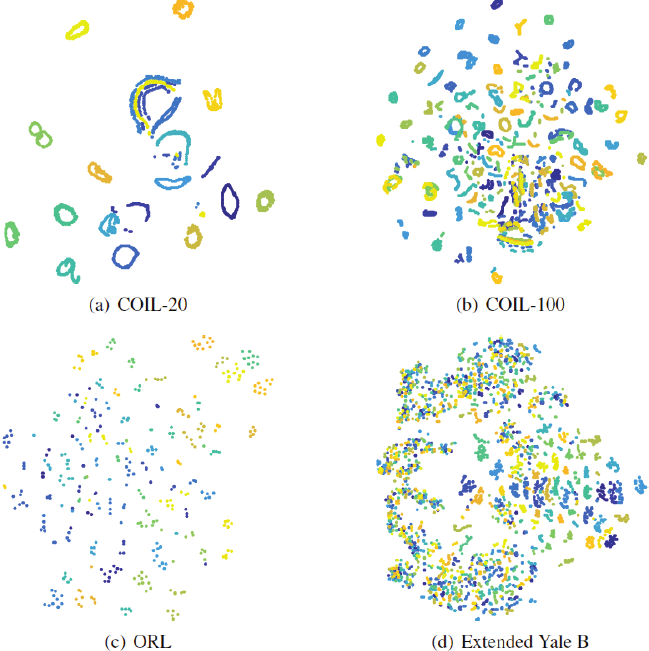}
\caption{Visualization of the learned feature with $t$-SNE.}\label{tsne}
\end{figure*}

The UDLL network architecture of COIL-20 consists of one-layer encoders and decoders with 15 channels of kernel size 3$\times$3 and a self-expressive layer with 1440 neurons. In the fine-tuning stage, we take all images as a single batch for training and we set regularization parameters to $\alpha = 1000$, $\beta = 1$, and $\gamma = 19$. The number of the fine-tuning epoch is set to 68. For COIL-20, the number of local nearest neighbors is set to $k = 3$ to construct the prior graph calculated by Eq.~\eqref{s_star}.

For COIL-100, the UDLL network architecture consists of one-layer encoder and decoder with 50 channels of kernel size 5$\times$5 and a self-expressive layer with 7200 neurons. In the fine-tuning stage, we take all images as a single batch for training and we set regularization parameters to $\alpha = 15$, $\beta = 1$, and $\gamma = 280$. The number of the fine-tuning epoch is set to 140. For COIL-100, the number of local nearest neighbors is $k = 5$ for the prior graph construction by Eq.~\eqref{s_star}.

For ORL, the UDLL network consists of three-layer encoder and decoder with $\{5, 3, 3\}$ channels of kernel sizes $\{5\times 5, 3\times 3, 3\times 3\}$ and a self-expressive layer with 400 neurons. In the fine-tuning stage, we take all images as a single batch for training and we set regularization parameters to $\alpha = 5, \beta = 1$, and $\gamma = 8$. The number of the fine-tuning epoch is set to 1550. For ORL, the number of local nearest neighbors assigned to each latent data point is $k = 3$ for the prior graph construction by Eq.~\eqref{s_star}.

Extended Yale B is larger than ORL. The UDLL network for Extended Yale B consists of three-layer encoder and decoder with $\{10, 20, 30\}$ channels of kernel sizes $\{5\times 5, 3\times 3, 3\times 3\}$ and a self-expressive layer with 2432 neurons. In the fine-tuning stage, we take all images as a single batch for training and we set regularization parameters to $\alpha = 3.2$, $\beta = 1$, and $\gamma = 0.01$. The number of the fine-tuning epoch is set to 1600. For Extended Yale B, the number of local nearest neighbors is $k=10$ for the prior graph construction by Eq.~\eqref{s_star}.

The Adam optimizer~\cite{kingma2014adam} is used to minimize the loss and the learning rate is set to 0.001 in all experiments.

\subsection{Experimental Results}
The learned features are visualized by $t$-SNE~\cite{maaten2008visualizing} as shown in Fig.~\ref{tsne}. It can be seen from Fig.~\ref{tsne} that UDLL separates latent features belonging to different classes very well and UDLL latent features have a distribution that are easily segmented. Fig.~\ref{tsne} implies the impressive representation ability of learned latent features $\Z$ by UDLL. Fig.~\ref{tsne}(d) does not well represent the latent feature since we vectorize a tensor concatenated by different channels of feature maps and vectorization renders it to lose two dimensional structure information.

For verifying the representation effectiveness of our UDLL, we test the learned features on clustering task. We evaluate the quantitative results through clustering accuracy (${\rm ACC}$) defined by,
\begin{equation}\label{acce}
  {\rm ACC}=\frac{\sum^n_{i=1}\delta(\tau_i,{\rm map}(r_i))}{n}
\end{equation}
where total $n$ data points are belonging to $k$ clusters, $\tau_i$ denotes the ground-truth label of the $i$-th sample, $r_i$ denotes the corresponding learned clustering label, and $\delta(\cdot,\cdot)$ denotes the Dirac delta function
\begin{equation}
  \delta(x,y)=
  \begin{cases}
  1\,,&{\rm if}~~x=y\,;\\
  0\,,&{\rm otherwise}
\end{cases}
\end{equation}
and map$(r_i)$ is the optimal mapping function that permutes the obtained labels to match the ground-truth labels. The best mapping is found by the Kuhn-Munkres algorithm~\cite{lovasz2009matching}.

\begin{table*}[!ht]
\centering
\caption{Clustering accuracy of different methods (ACC \%).}\label{result}

\begin{tabular}{c|cccccccccc}
  \hline
   &LRR&LRSC&SSC&KSSC&EDSC&SSC-OMP&DEC&DSC&UDLL\\
  \hline
  COIL-20      &68.99&68.75&85.14&75.35&85.14&54.10&79.00&94.86&97.57\\
  COIL-100     &40.18&49.33&55.00&52.82&61.87&33.61&60.66&69.04&70.86\\
  ORL          &61.75&67.50&67.70&65.75&72.75&64.00&60.33&86.00&87.75\\
  Extend Yale B&65.13&70.11&72.49&72.25&88.36&75.29&48.66&97.33&97.70\\
\hline
\end{tabular}
\end{table*}

We conduct experiments on four datasets to demonstrate the effectiveness of our feature representation ability of the proposed UDLL algorithm. We compare our algorithm with several baselines:
\begin{itemize}
  \item Low-rank representation (LRR)~\cite{liu2013robust} solved subspace clustering problem by seeking the lowest rank representation among all candidates that can represent the data samples as linear combinations of bases in a given dictionary.
  \item Low-rank subspace clustering (LRSC)~\cite{Vidal2014} posed the subspace clustering problem as a non-convex optimization problem whose solution provides an affinity matrix for spectral clustering, and the goal was to decompose the corrupted data matrix as the sum of a clean and self-expressive dictionary plus a matrix of noise/outliers or gross errors.
  \item Sparse subspace clustering (SSC)~\cite{elhamifar2013sparse} aimed to find a sparse representation among the infinitely many possible representations of a data point in terms of other points.
  \item Kernel sparse subspace clustering (KSSC)~\cite{patel2014kernel} extended SSC to nonlinear manifolds by using the kernel trick.
  \item Efficient dense subspace clustering (EDSC)~\cite{Ji2014} dealt with subspace clustering by estimating dense connections between the points lying in the same subspace and formulated subspace clustering as a Frobenius norm minimization problem.
  \item SSC by orthogonal matching pursuit (SSC-OMP)~\cite{You2016} proposed a subspace clustering method based on orthogonal matching pursuit that is computationally efficient and guaranteed to provide the correct clustering.
  \item Deep Embedded Clustering(DEC) conducted unsupervised clustering by first pretraining a deep autoencoder, and then fine-tuning the autoencoder to perform deep embedding learning and clustering jointly.
  \item Deep subspace clustering networks (DSC)~\cite{ji2017deep} was based on a deep autoencoder to find the coefficient representation matrix and applied it to spectral clustering to obtain the clustering results.
\end{itemize}

The source code of these baselines released by the authors are used and we tune the parameters by grid search to achieve the best results on each datasets.

In the clustering experiments, once we obtain the new affinity graph $\bm{W}$, we can use it to construct an affinity matrix for spectral clustering as same as used in most existing methods. Although affinity matrix $\frac{\bm{W} + \bm{W}^\T}{2}$ can be directly fed to spectral clustering, many heuristics have been developed to improve the performance of the constructed affinity matrix. In this paper, we utilize the heuristics used by EDSC~\cite{Ji2014}.

The clustering accuracy of different algorithms on all datasets is provided in Table.~\ref{result}.

\begin{table}[ht]
\centering
\caption{Clustering accuracy of different features (ACC \%).}\label{as}
\begin{tabular}{c|ccccccc|ccc}
  \hline
   &PT$+$SSC&PT$+$EDSC&UDLL\\
  \hline
  COIL-20      &77.92&84.21&97.57\\
  COIL-100     &56.07&61.12&70.86\\
  ORL          &73.25&73.75&87.75\\
  Extend Yale B&74.67&87.36&97.70\\
\hline
\end{tabular}
\end{table}
It can be seen from Table.~\ref{result} that our UDLL algorithm outperforms all of the state-of-the-art methods, which greatly validate the effectiveness of the locality-preserving loss. From this result, we can make the following conclusions:
\begin{itemize}
    \item The clustering accuracy of DSC and UDLL have a overwhelming advantage over the rest of methods, which can demonstrate the great representation ability of the DNNs as well as the powerful architecture of the autoencoders.
    \item The UDLL outperforms the DSC in terms of accuracy on all of the four datasets, implying the good quality of the refined affinity matrix for more accurate clustering result.
    \item The locality-preserving loss can well capture the local connections from pretrained GNN feature space to the fine-tuned stage feature space, improving the representation ability of affinity matrix.
    \item We formulate our novel local-preserving loss under a strong theoretical foundation of Ncut, and the intra-class compactness and inter-class separation is the key to improve the clustering accuracy.
\end{itemize}
\subsection{Ablation Study}
\begin{figure}[ht]
\centering
\includegraphics[width=2.2in]{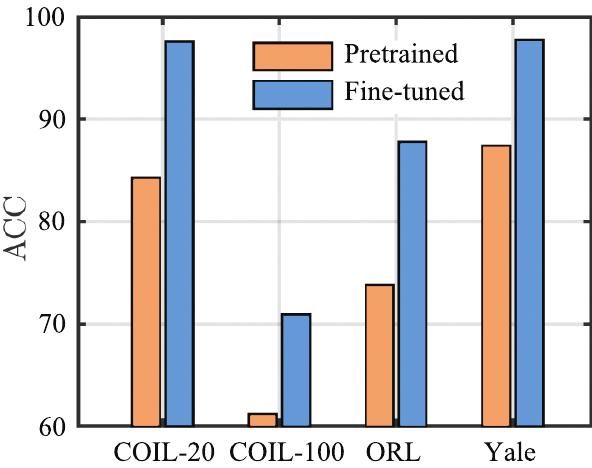}
\caption{Representation ability of pretrained and fine-tuned latent feature (ACC \%). }\label{comp}
\end{figure}

To further evaluate the feature representation ability of UDLL and the effectiveness of the structure connection constraint locality-preserving loss, we directly apply EDSC to the pretrained graph and the fine-tuned graph, respectively. As shown in Fig.~\ref{comp}, the fine-tuned feature has better performance than the pretrained feature.

Comparing with the UDLL clustering results with fine-tuned features, we show the different results using pretrained features with SSC (PT$+$SSC) and EDSC (PT$+$EDSC) in Table~\ref{as}. It can be seen from Table~\ref{as} that the fine-tuned model obtains the better results than others.
\section{Conclusion}\label{cld}
We proposed a new algorithm called UDLL. UDLL is a generative model with an additional self-expressive layer. The self-expressive layer is used to compute the coefficient matrix and the matrix is used to construct an affinity matrix for post-processing spectral clustering. First, we pretrained autoencoders and obtain the latent representation of input data. Second, the latent representation was used to construct a prior graph which describes the affinity between pairwise latent features. Third, we fine-tuned the UDLL model with a self-expressive layer and with connectivity regularization by the prior graph. The prior graph was constructed by a $k$-nearest neighbors algorithm. The prior graph formed the data structure of pretrained latent feature which was preserved from pretraining to fine-tuning by optimizing the locality-preserving loss. By considering the connectivity constraint in the prior graph, experiments on four image datasets had demonstrated that UDLL feature has a powerful representation ability and UDLL provided a significant improvement over state-of-the-art methods.
{\small
\bibliographystyle{ieee}

}
%
%
%
\end{document}